 \newcommand{\ARXIV}{} 

\ifdefined\ARXIV
  \documentclass[letterpaper, 10pt, twocolumn]{article}
  \usepackage[margin=0.75in]{geometry}
  \usepackage{libertine}
  \renewcommand\footnotemark{} 
  \date{}
\else
  \newcommand{\CLASSINPUTtoptextmargin}{17.6mm}
  \newcommand{\CLASSINPUTbottomtextmargin}{25.4mm}
  \newcommand{\CLASSINPUTinnersidemargin}{16.2mm}
  \newcommand{\CLASSINPUToutersidemargin}{16.2mm}
  \documentclass[letterpaper, conference]{IEEEtran}
  \IEEEoverridecommandlockouts
  \let\labelindent\relax 
\fi

\usepackage{amsmath,amssymb,amsfonts}
\usepackage{algorithmic}
\usepackage{cuted} 
\usepackage{enumitem}
\usepackage{textcomp}
\usepackage[thinlines]{easytable}

\usepackage{graphicx}
\usepackage{subfig}
\usepackage{xcolor}
\usepackage{hyperref}
\hypersetup{colorlinks = true, citecolor = {black}, linkcolor = black}

\begin{document}


\title{\LARGE \bf Flexure-based Environmental Compliance \\ for High-speed Robotic Contact Tasks
 \thanks{Authors are with the Department of Automation at Fraunhofer IPK, Berlin, Germany. Emails: \texttt{\{richard.hartisch,\, kevin.haninger\}@ipk.fraunhofer.de}}
 \thanks{This project has received funding from the European Union's Horizon 2020 research and innovation programme under grant agreement No  820689 — SHERLOCK.}
}

\author{Richard Hartisch and Kevin Haninger}

\maketitle

\begin{abstract}
  The design of physical compliance -- its location, degree, and structure -- affects robot performance and robustness in contact-rich tasks.  While compliance is often used in the robot's joints, flange, or end-effector, this paper proposes compliant structures in the environment, allowing safe and robust contact while keeping the higher motion control bandwidth and precision of high impedance robots. Compliance is here realized with flexures and viscoelastic materials, which are integrated to several mechanisms to offer structured compliance, such as a remote center of compliance. Additive manufacturing with fused deposition modeling is used, allowing faster design iteration and low-cost integration with standard industrial equipment. Mechanical properties, including the total stiffness matrix, stiffness ratio, and rotational precision, are analytically determined and compared to experimental results. Three remote center of compliance (RCC) devices and a 1-DOF linear device are prototyped and tested in high-speed assembly tasks. 
\end{abstract}


\section{Introduction}
Physical compliance enables safe contact-rich tasks on robots: reducing contact forces \cite{bicchi2004}, improving contact stability \cite{calanca2016}, and often improving task robustness \cite{yun2008}. Compliance is typically introduced in either the joints (as series-elastic actuators \cite{pratt1995} or torque sensors \cite{albu-schaffer2007}), the end-effector \cite{shintake2018}, or the flange \cite{drake1978}.  

However, compliance involves design trade-offs. Reducing stiffness limits feedback motion control gains, reducing motion control bandwidth \cite{bicchi2004}. Compliance also affects sensing, reducing the ability to detect changes in the environment dynamics \cite{haninger2018} and sensing efficacy \cite{kashiri2017}. Compliance also typically reduces the maximum payload - currently, the highest payload for joint-torque-controlled commercial robots is 14kg (KUKA iiwa). 

We propose integrating compliance to the environment, noting that most robots operate in human-built environments (except field robotics). If robots {\it must} operate in environments designed for humans, robots will likely require human-like intrinsic dynamics. However, if the environment can be adapted -- e.g. made compliant -- high-impedance robots could be used, keeping their benefits of higher payload, motion control bandwidth, and accuracy.

This approach has the disadvantage of requiring additional design and integration work in the environment - e.g. assembly fixtures or table. This additional design work can be made easier with solutions which are (i) easily adjusted to allow in-situ tuning and (ii) easily integrated into industrial environments. Towards this, this paper presents the use of additive manufacturing, precisely fused deposition modeling, also known as fused filament fabrication, for flexure joints which can be integrated to compliant tables, fixtures, feeders, and storage magazines. In addition to reducing stiffness, we seek to reduce the sprung inertia, so that collision forces and high-speed inertial effects are reduced, towards improving robustness and reducing collision danger to robots, environments, and humans.

\begin{figure}[t!]
	\centering
	\subfloat[Small RCC fixture \label{small_scale}]{\includegraphics[width=.45\columnwidth]{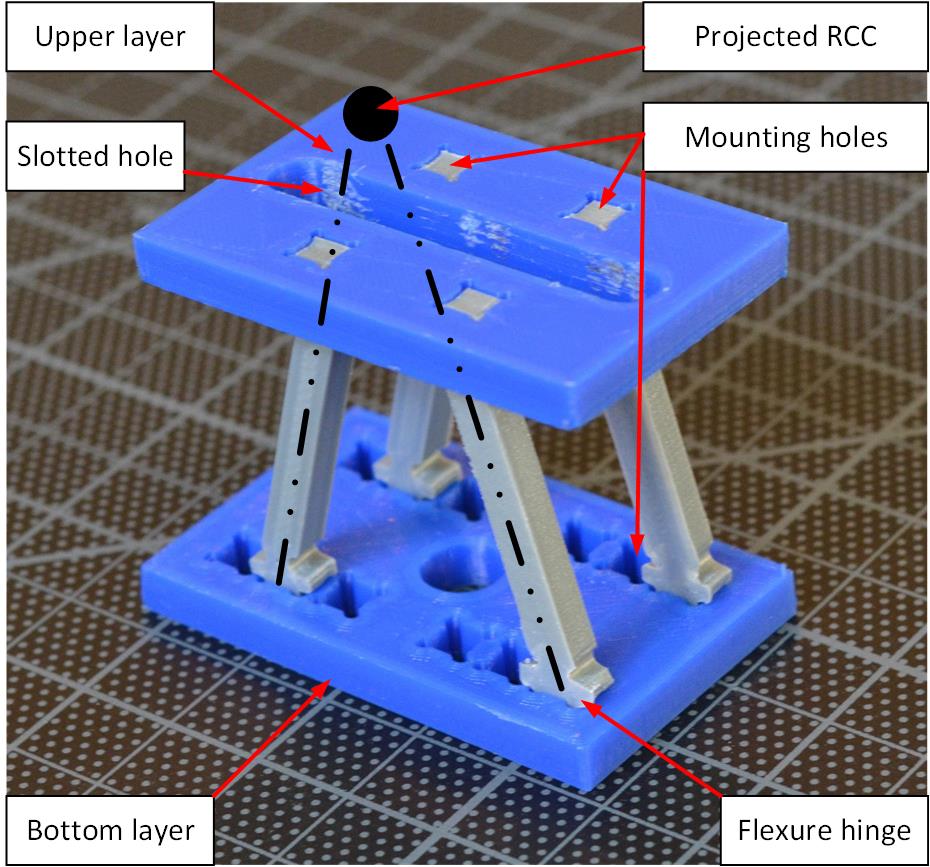}} 
	\hfill
	\subfloat[Medium scale RCC and vertical compliance table 
	\label{large_scale}]{\includegraphics[width=.5\columnwidth]{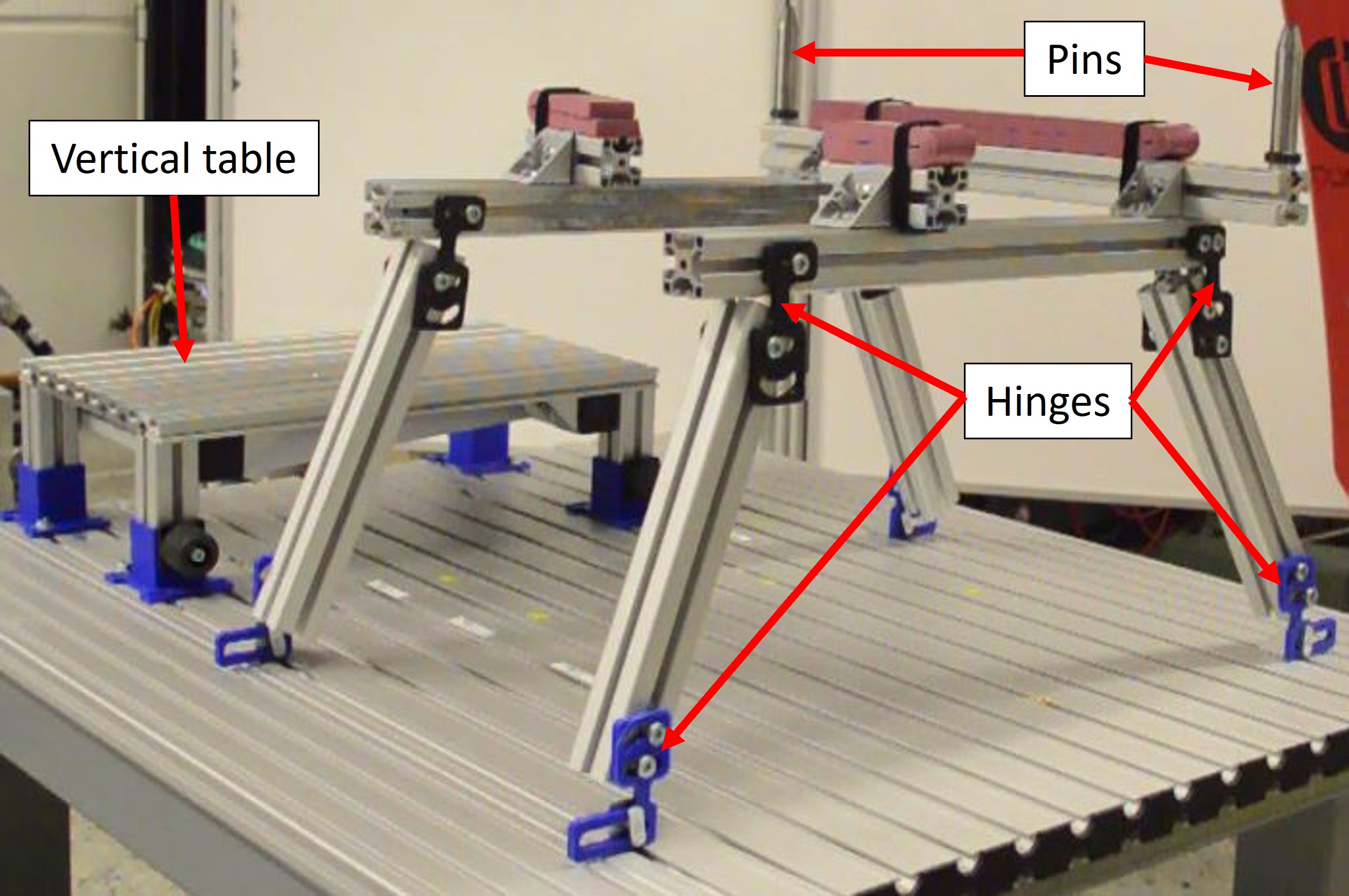}}
	\protect\caption{Flexure-based compliant mechanisms realized with (a) additive manufacturing of compliant legs (gray) and (b) compliant joints (black, blue) and aluminum profile.}
\end{figure}

Flexure joints can replace standard revolute joints, allowing for movement between rigid bodies via elastic deformation of a narrowing in a single piece joint. Flexures are commonly used in precision engineering and microtechnology with high demands on positioning accuracy \cite{linss2015}, where they are often manufactured by etching. Flexures can also be produced through additive manufacturing, which is cost-effective and can reduce the part count. Nonetheless, the angle of deflection is restricted, limiting the range of motion of the mechanism \cite{linss2017, lai2016}. However, recent designs of flexure joints allow for wider range of movements while still being additively manufactured \cite{rommers2021}. Mechatronic design of flexure mechanisms with actuators for planar micro-positioning have also been studied \cite{yong2012, kang2005}.

Here, flexures are used for larger passive compliant mechanisms, realizing a remote center of compliance (RCC) \cite{drake1978}. Currently, RCCs are standard for peg-in-hole assembly, and the RCC compensates lateral and orientation misalignment \cite{whitney2004}. Usually, for these types of applications an RCC is mounted directly on the robot flange, where the RCC is realized by laminated shear pads between two plates \cite{ati}, while pneumatic \cite{bottero2020} and flexible beam \cite{lai2016} approaches have been proposed.  Flexure-based RCCs in the robot end-effector have been proposed in design studies \cite{ciblak2003, havlik2011}, but not implemented. In addition to RCC devices, we present a linear compliant mechanism based on viscoelastic materials.

Determining what compliance is needed is a difficult design problem \cite{gallego2009}, with proposed methods including error-correcting motions \cite{schimmels1993} or freedom and constraint methods \cite{hopkins2010}. However, these techniques are difficult to scale to 6-DOF and for assemblies with complex geometries. For compliant mechanism design objectives such as kinematic workspace, natural frequency, and stiffness ratio, topological \cite{jin2018a} or parametric \cite{wang2017b} optimization have been proposed. Typically, compliance is optimized with application-specific heuristics, such as in drilling \cite{bu2017}, peg-in-hole assembly \cite{yun2008}, or deburring \cite{schimmels2001}.  

Our contribution is the design, analysis and validation of compliant mechanisms for high-speed robot assembly. Compliance is well-established in robots \cite{pratt1995, albu-schaffer2007, shintake2018}, and used informally in the environment for lab experiments \cite{katsura2007, ferraguti2019}; we develop the first known application of structured compliant mechanisms in the environment. Flexure-based RCC devices in the robot flange or end-effector have been proposed \cite{havlik2011}, but are here physically validated. A design study for a 3-DOF RCC in the environment has been done \cite{jiang2013}, here a 6-DOF mechanism  is studied and physically prototyped. First the design and validation of a single flexure is introduced, then an RCC mechanism built from several flexures is presented and the 6-DOF stiffness matrix calculated. This stiffness is experimentally validated, and several applications at small and large scale are presented, with experiment videos available here: \url{https://owncloud.fraunhofer.de/index.php/s/l7ZY2MpXQhRTUq2}

\section{Flexure Design}
This section presents the designs of individual hinge joints and validates their mechanical properties.

\subsection{Flexure Hinge Design}
Flexures use local narrowing of a structural member to induce local bending. This provides a joint which, compared with classical kinematic joints, introduces no mechanical play and has no sliding of bearing surfaces.  On the other hand, flexures have a limited range of motion and generally have a lower stiffness in all directions.

\begin{figure}
    \centering
    \subfloat[Small flexures with varying degree offsets
	\label{small_scale_columns}]{\includegraphics[width=.48\columnwidth]{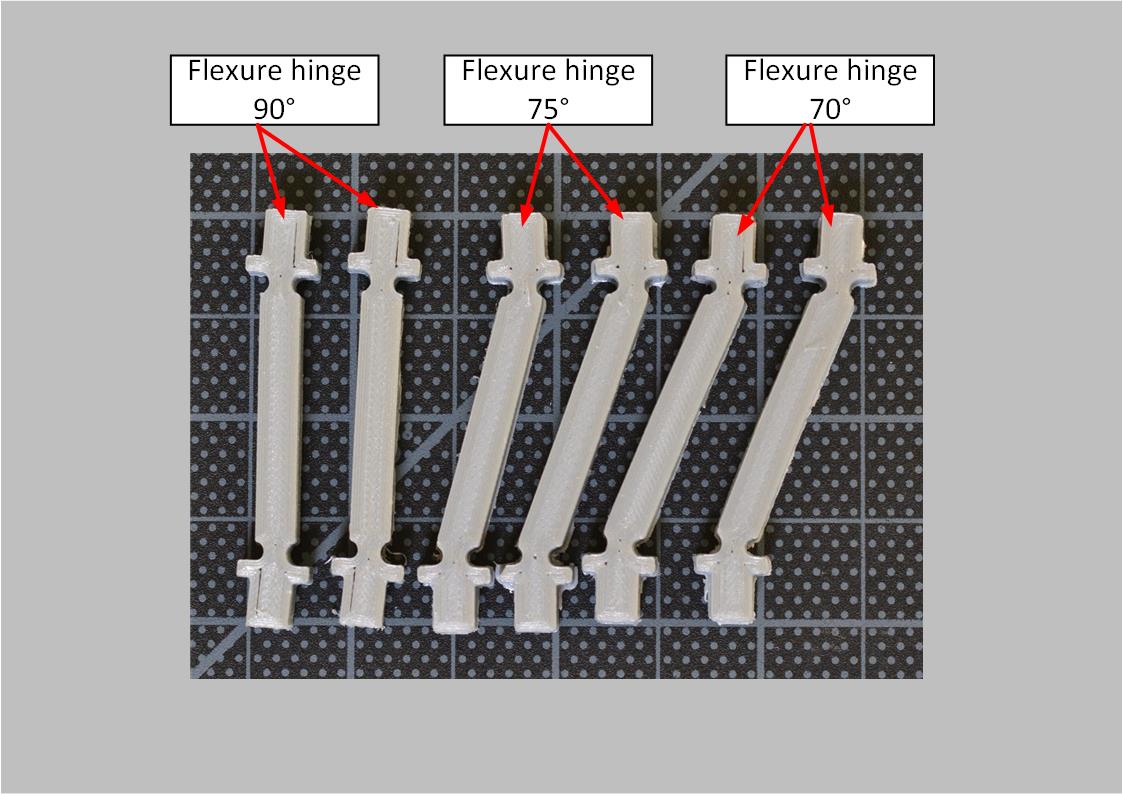}}
	\hfill
    \subfloat[Individual hinges from PLA
	\label{large_scale_versions}]{\includegraphics[width=.48\columnwidth]{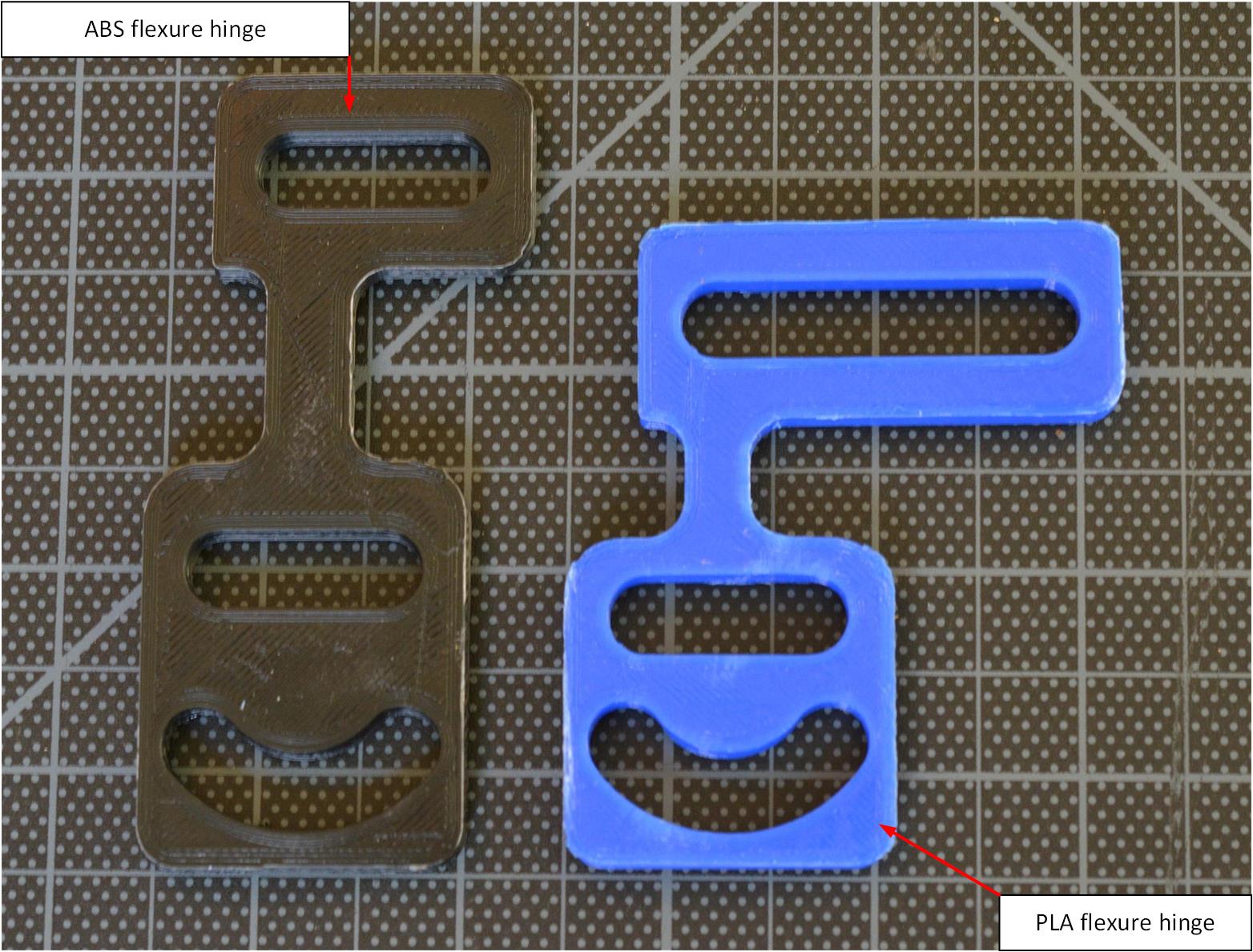}}
    \caption{Individual flexures, (a) with integrated two hinge joints and connecting column, and (b) individual hinge joint for integration with larger structures}
    \label{flexures}
\end{figure}

The proposed flexures are shown in Figure \ref{flexures}.  For smaller flexures seen in Figure \ref{small_scale_columns}, two hinge joints are integrated with a connecting column of width/thickness $5$ mm where the hinge joints are realized with circular cutouts of radius $1.25$ mm.  For the larger hinge joints in \ref{large_scale_versions}, a simple rectangular narrowing of a width $10-17.5$ mm and thickness $6-7$ mm is used, along with elongated holes to allow for flexibility in orientation and location of mounting.

\subsection{Manufacturing and Materials}
The flexures here are additively manufactured with fused deposition modeling, using a flexible material (DSM Arnitel Eco, Innofill Innoflex 45) for the small flexures in Figure \ref{small_scale_columns}, while ABS (Verbatim ABS), PLA, or Tough PLA (Ultimaker Tough PLA) are used for the larger hinge joints in Figure \ref{large_scale}, \ref{large_scale_versions} and \ref{single_magazine}.

Cura was used for slicing, with a layer height of 0.15 mm and 20\% infill, except for the hinges used in Figure \ref{single_magazine}, where gradual infill was chosen to withstand higher loads. 

\begin{figure}[h!]
	\centering
	\hfill
	\hfill
	\subfloat[Components of linear system
	\label{z_direction}]{\includegraphics[width=.533\columnwidth]{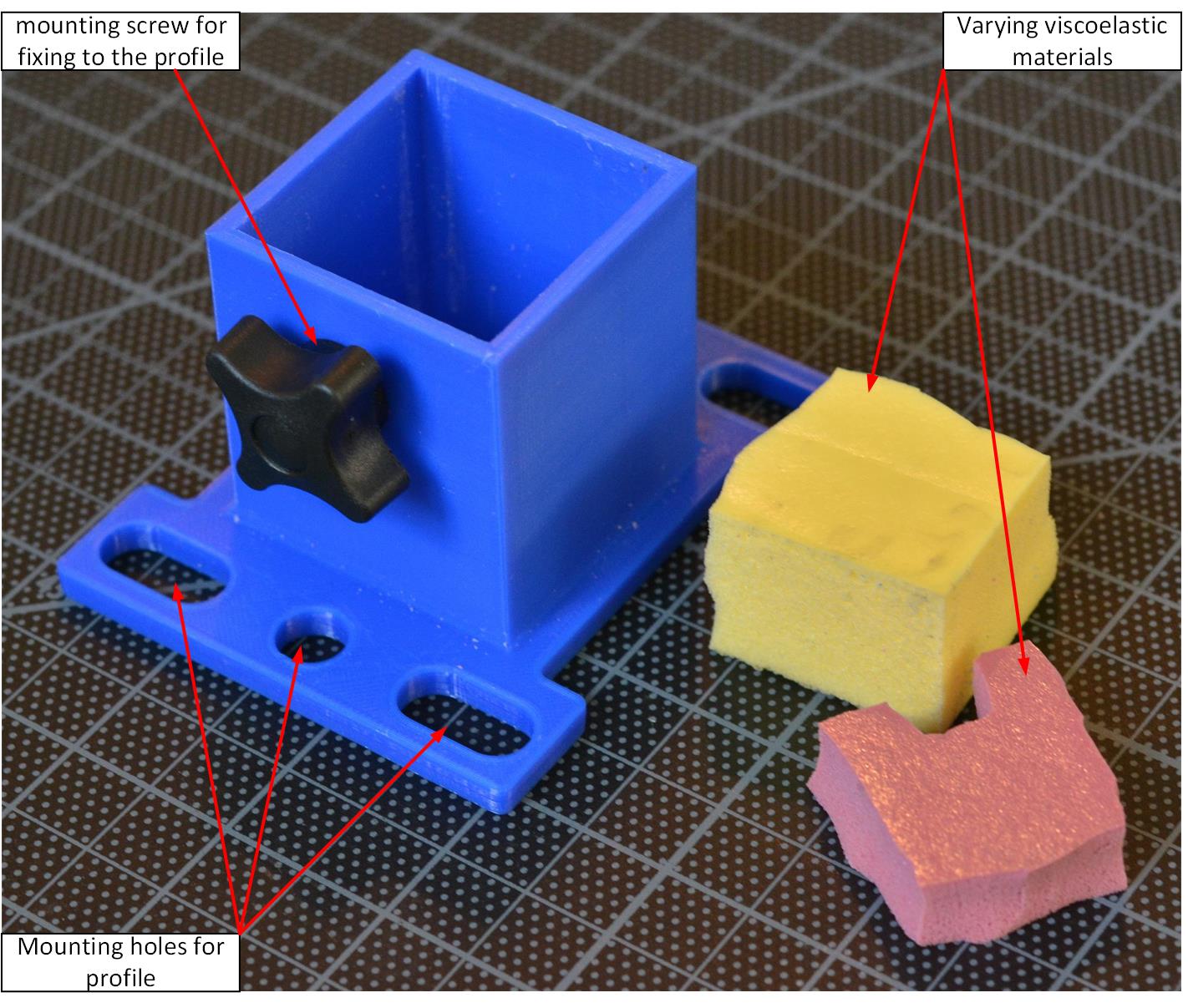}}
	\hfill
	\subfloat[Linear assembly
	\label{z_direction_assembly}]{\includegraphics[width=.34\columnwidth]{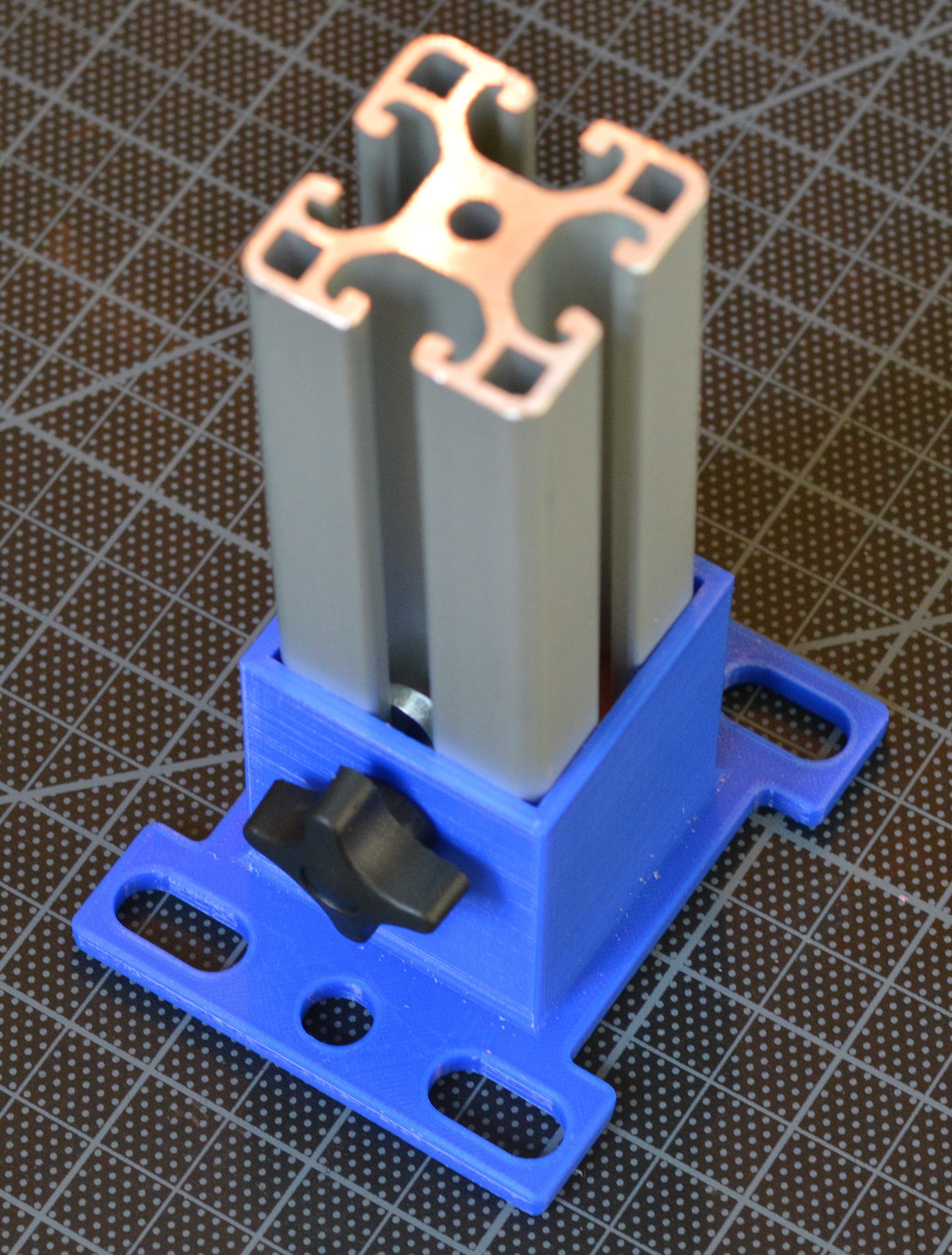}}
	\hfill
	\hfill
	\protect\caption{In (a), the construction of vertical compliance is displayed with different viscoelastic materials. In (b), assembled with standard profile, with a hand screw which can lock vertical compliance}
\end{figure}

Viscoelastic material (Sylomer SR 11 - 25 and SR 42 - 12) as seen in Figure \ref{z_direction} is used to realize a linear compliant mechanism, where a box profile constrains the other DOFs as seen in Figure \ref{z_direction_assembly}. Mounting holes are used to mount the box profile on aluminum construction profiles, and a hand screw allows the vertical compliance to be locked. 

\subsection{Single Joint Mechanical Properties}
The stiffness and maximum load were tested on four different configurations regarding the flexure material and width/thickness of the flexure bridge. Two matching joints have been mounted on either side of $40\times40$mm aluminum profile to form a rotational joint. A Comau Racer with a Force/Torque sensor applied forces at a $200$ mm distance, in both the joint direction (longitudinal) and cross from the joint direction. The results can be seen in \ref{tab:mech_prop}, where a stiffness ratio of $7.6$ and $8.1$ is achieved for the $6$ mm thickness joints.

Additionally, creep behavior of the large Tough PLA flexure used in \ref{single_magazine} has been tested by applying a load of 22 N then measuring forces from a fixed robot position.  Creep resulted in an exponential decay of the force, with steady state force of 19 N and a time constant of $200$ seconds.    

\begin{table}[h!]
\centering
\begin{tabular}{|l||c|c|c|}
\hline
Flexure variant & Cross Stiff & Joint Stiff & Max Jt Load\\
\hline
\hline
PLA narrow 6mm  & - & 66.412 & 6.25\\
\hline
ABS narrow 6mm  & 604.963 & 75.0225 & 8.625\\
\hline
TPLA narrow 6mm & 564.9776  & 74.635 & 10.5\\
\hline
TPLA wide 7mm & 974.969 & 280.129 & 22.95\\
\hline

\end{tabular}
\caption{Mechanical properties per two-joint hinge, stiffness in Nm/rad, Max Joint Load in Nm.}
\label{tab:mech_prop}
\end{table}



\section{Compound Flexure Mechanism} 
This section presents the composition of multiple flexures and analytic calculation of the resulting $6\times6$ compliance matrix. By employing multiple flexures, 4-bar and other mechanisms can be constructed with a rotational stiffness in the joints. The geometry of this mechanism can be designed to achieve directional translational compliance or rotation about a remote center as an RCC.

This principle has been used to realize compliance of a small RCC fixture \ref{small_scale}. The upper layer is printed with a slotted hole to ensure varying mounting positions for applications while the lower layer has multiple leg mounting holes to allow adjustment to the location of the center of compliance, which is projected in Figure \ref{small_scale}. The same principle is projected onto larger mechanisms, where two hinges are attached at each end of a standard aluminum profile, as seen in Figures \ref{large_scale_versions} and \ref{single_magazine}.

\subsection{Stiffness Matrix of RCC Device}
To calculate the total stiffness matrix of the small RCC device, the compliance matrix of the hinges and the rectangular beam connecting the hinges is calculated, following \cite{tuo2020}, \cite{pham2005}, and \cite{lai2016}. In the following, it is assumed that deformations are small to ignore nonlinear effects \cite{tuo2020}. 

To calculate the stiffness matrix for the small RCC, the Young's modulus $E$ is taken from the material datasheet, but the shear modulus $G$ is not specified. Hence, it can be determined using the assumption that the printed part is isotropic as $ G  = \frac{E}{2(1+\nu)} $, with Poisson's ratio $\nu$. Since there is no published statement regarding the Poisson's ratio for the used material, a Poisson's ratio of $\nu = 0.48$ is assumed from thermoplastic copolyester.

The units for elements of the stiffness matrix are: ($K_{11}$, $K_{22}$, $K_{33}$) are N/mm, for ($K_{26}$ and $K_{35}$) N/rad, Nmm/mm for ($K_{53}$ and $K_{62}$) and Nmm/rad for ($K_{44}$, $K_{55}$ and $K_{66}$). The compliance matrix is defined as the inverse of the stiffness matrix $C = K^{-1}$, with resulting units.

\subsubsection{Compliance Matrix of the Rectangular Beam}
To calculate the compliance matrix of the rectangular beam, the torsional moment of inertia can be calculated as $I_t = \beta h w^3$, with the torsional shape coefficient $\beta$, the height $h$ and the width $w$ of the rectangular cross section. To calculate the torsional shape coefficient $\beta$, the length-width ratio of the rectangular cross section is needed, which leads to the torsional shape coefficient $\beta_y$, using equation (8), or Table 1 from \cite{tuo2020}.
This leads to a torsional moment of inertia $I_t$ of $88.021$ mm$^4$. For the shear coefficient $\alpha = 6/5$ is assumed for a rectangular cross section. Using the length $l$ and the intermediate value $s$, calculated as $2r + t$, with the thickness $t$ at the thinnest part of the notch, given as 2.82 mm and the radius being $r = 1.25$ mm, the compliance matrix is 
\begin{equation*}
C_{beam} = \\
\begin{bmatrix}

\frac{l}{Ews} & 0 & 0 & 0 & 0 & 0 \\
0 & C_{22} & 0 & 0 & 0 &\frac{6l^2}{Ews^3} \\
0 & 0 & C_{33} & 0 & \mathtt{-}\frac{6l^2}{Ew^3s} & 0 \\
0 & 0 & 0 & \frac{l}{GI_{t}} & 0 & 0 \\
0 & 0 & \mathtt{-}\frac{6l^2}{Ew^3s} & 0 & \frac{12l}{Ew^3s} & 0 \\
0 & \frac{6l^2}{Ews^3} & 0 & 0 & 0 & \frac{12l}{Ews^3} 
\end{bmatrix}
\end{equation*}
where $C_{22} = \frac{\alpha l}{Gws} \mathtt{+} \frac{4l^3}{Ews^3}$ and $C_{33} = \frac{\alpha l}{Gws}\mathtt{+}\frac{4l^3}{Ew^3s}$ \cite{tuo2020}.

%

\subsubsection{Compliance Matrix of the Hinge}
The compliance matrix of a single hinge is calculated using equations (12)-(16) from \cite{tuo2020}. The torsional moment of inertia has to be calculated for an infinitesimal strip $dx$ at a position $x$ along the $x$ direction of the notch (as shown in Figure \ref{rcc_modelling}) and therefore varies along the circular hinge. This is necessary to calculate $C_{\theta_{x}-M_{x}}$, for which a case distinction has to be made. The thickness $t$ at the thinnest part of the notch is given as 2.82 mm, width $w=5$ mm, and radius $r=1.25$ mm. According to Table 3 of \cite{tuo2020}, condition 2 applies to the $C_{\theta_{x}-M_{x}}$ element of the resulting compliance matrix \eqref{flexure_stiffness_matrix}, where $(f_1, f_2, f_3, f_4)$ are calculated according to (13)-(16) of \cite{tuo2020}.

\begin{figure*}[!t]
\vspace{1mm}
\begin{equation}
C_{hinge} = 
\begin{bmatrix}
\label{flexure_stiffness_matrix}
\frac{l}{Ew}f_1 & 0 & 0 & 0 & 0 & 0 \\
0 & \frac{\alpha l}{Gw} f_1 + \frac{12}{Ew} f_1 & 0 & 0 & 0 &\frac{12 (r+h_1)}{Ew} f_4 \\
0 & 0 & \frac{\alpha}{Gw} f_1 +\frac{12}{Ew^3} f_3 & 0 & -\frac{12(r+h_1)}{Ew^3} f_1 & 0 \\
0 & 0 & 0 & C_{\theta_{x}-M_{x}} & 0 & 0 \\
0 & 0 & -\frac{12(r+h_1)}{Ew^3} f_1 & 0 & \frac{12}{Ew^3} f_1 & 0 \\
0 & \frac{12(r+h_1)}{Ew} f_4 & 0 & 0 & 0 & \frac{12}{Ew} f_4 
\end{bmatrix}
\end{equation}
\end{figure*}

\subsubsection{Compliance Matrix of a Flexure}
\begin{figure}[h!]
	\centering
	\includegraphics[width=.75\columnwidth]{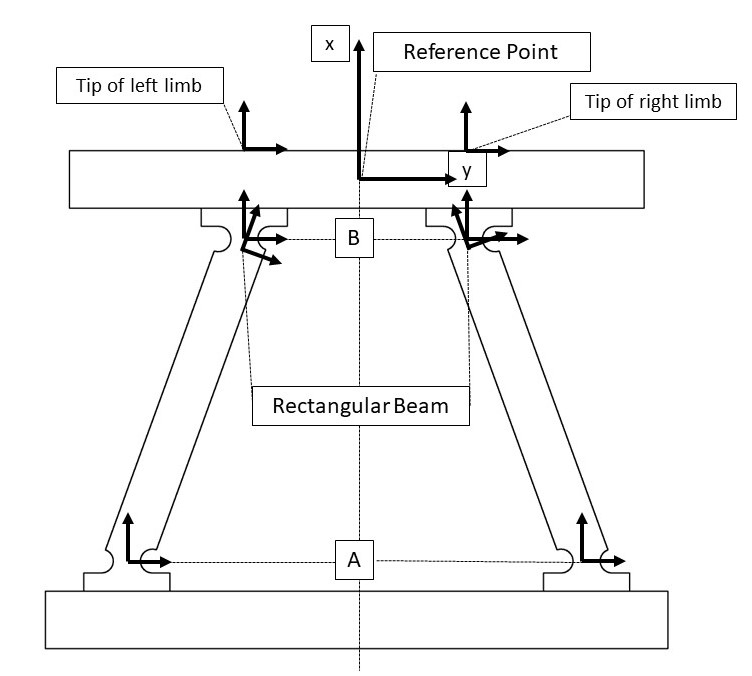}
	\hfill
	\protect\caption{Modelling of a planar RCC, where the flexure members and its corresponding coordinate systems are illustrated \label{rcc_diagram} \label{rcc_modelling}}
\end{figure}

A complete flexure, or limb, is a serial chain of two hinges and a connecting rectangular beam. To calculate the effective total stiffness $K_{i}$ of a flexure, the amplification matrix of the displacement for each member of the serial chain has to be determined. This uses the transformation matrix from the reference point to the tip of the limb, denoted $J_i = J_{Fi}^{-T}$ \cite{pham2005}, where $-T$ is the inverse transpose. $J_{Fi}$ is

\begin{equation}
\begin{aligned} 
J_{Fi} & = \begin{bmatrix}
R_z(\theta_i) & \textbf{0}_{3,3} \\
S(r_i)R_z(\theta_i) & R_z(\theta_i) 
\end{bmatrix},\\
S(r_i) & = \begin{bmatrix}
0 & r_{iz} & -r_{iy} \\
- r_{iz} & 0 & r_{ix} \\
r_{iy} & -r_{ix} & 0  
\end{bmatrix},
\label{amplification_matrix}
\end{aligned}
\end{equation}
where $R_z(\theta)$ is the standard $3\times3$ rotation matrix about $z$ with $\theta$ denoting the rotation around the $z$ axis, and the entries $r_{ix}$, $r_{iy}$ and $r_{iz}$ are components of the vector $r_{i}$, pointing from each member to the tip of the $i$th limb, starting from each coordinate system, as shown in Figure \ref{rcc_modelling} \cite{lai2016}. 

Each member and tip with its corresponding coordinate system is shown in Figure \ref{rcc_modelling}. For these flexures $r_{iz}$ is zero, since the tip is in the same x-y plane as each member of the serial chain. The displacements for $r_{ix}$ and $r_{iz}$ can be found in table \ref{tab:displacements}. Since the rectangular beam is oriented with an angle of 70° to the lower plane an assumption has to be made regarding the values for $r_{beam,x}$, $r_{beam,y}$. Both displacements are measured from the top of the rectangular beam.

\begin{table}[h!]
\centering
\begin{tabular}{|l||c|c|c|c|}
\hline
Element & $r_{ix}$ [mm] & $r_{iy}$ [mm] & $\theta_i$ [deg] \\
\hline
\hline
Left Hinge A  & 42.85 & 14.765 & 0 \\
\hline
Left Rectangular Beam  & 10.4 & 0 & 20 \\
\hline
Left Hinge B & 9.15  & 0 & 0 \\
\hline
Right Hinge A  & 42.85 & -14.765 & 0  \\
\hline
Right Rectangular Beam  & 10.4 & 0 & -20\\
\hline
Right Hinge B  & 9.15  & 0 & 0 \\
\hline
\hline
\end{tabular}
\caption{Displacement components of chain members}
\label{tab:displacements}
\end{table}


After obtaining the amplification matrices of each member to the tip, the compliance matrix of the flexures can be calculated using the calculated compliance matrices $C_{i}$ and the amplification matrices $J_{i}$ of each member as \cite{pham2005, lai2016}
\begin{align}
& C_{limb} = \sum J_{i} C_{i}J_{i}^T.  
\end{align}

\subsubsection{Stiffness of Complete Mechanism}
Now, since the compliance matrix measured from the tip of the flexure is known, a connection to the reference point is needed. The amplification matrix of the displacement from the tip of each flexure to the reference point according to \cite{pham2005} can be determined analogous to the last section, but now $r_{iz}$ has is nonzero, measuring the middle of the tip of the flexure to the reference point in the middle of the upper platform. The displacements are listed in \ref{tab:displacements_tip}. 

\begin{table}[h!]
\centering
\begin{tabular}{|l||c|c|c|}
\hline
Limb & $r_{ix}$ [mm] & $r_{iy}$ [mm] & $r_{iz}$ [mm] \\
\hline
\hline
Limb left back  & -2.5 & 10.325 & -8.65\\
\hline
Limb right back  & -2.5 & -10.325 & -8.65\\
\hline
Limb left front  & -2.5 & 10.325 & 8.65\\
\hline
Limb right front  & -2.5 & -10.325 & 8.65\\
\hline
\hline
\end{tabular}
\caption{Displacement of limbs}
\label{tab:displacements_tip}
\end{table}

The stiffness matrix of a parallel combination, denoted as a linear spring $ls$, can now be calculated using the stiffness, transformation and amplification matrix of each limb, as \cite{pham2005} 
\begin{align}
& K_{ls} = \sum J_{F, limb,i} \cdot K_{limb, i} \cdot J_{limb, i}^{-1}  
\end{align}



The determined stiffness matrix for the total system is then
\begin{equation}
K = \begin{bmatrix}

0.16 & 0 & 0 & 0 & 0 & 0 \\
0 & 0.030 & 0 & 0 & 0 & 0.36 \\
0 & 0 & 0.0024 & 0 & 0.052 & 0 \\
0 & 0 & 0 & 4.7 & 0 & 0 \\
0 & 0 & 0.052 & 0 & 21.9 & 0 \\
0 & 0.36 & 0 & 0 & 0 & 8.6
\end{bmatrix} \cdot 10^3.
\end{equation}

The center of compliance can be derived as $C_{3,3}/C_{5,3}$ \cite{lai2016}, the center of the $z$ rotation is at $26.6$ mm above the reference point. The expected center of rotation given by idealized geometry, i.e. if the system were treated as an ideal 4-bar linkage as seen in Figure \ref{small_scale}, is $28.6$ mm. This gives a rotational precision of $2.0$ mm.

\subsection{Experimental Validation}

In this subsection the analytically determined stiffness matrix is compared to the experimentally determined directional stiffness of the RCC. To determine the directional stiffness experimentally, the flexures have been glued to the lower and upper platform. For the $y$-direction (according to the convention used in \ref{rcc_modelling}) the experimentally determined stiffness is $8,3-10$ N/mm, the stiffness in $z$-direction is $2.54$ N/mm.

Comparing the experimentally determined values with the analytical values shows, that the deviation is by $\approx$  5\% for the z-direction and $\approx$ 68\% - 74\% for the y-direction.
The assumption that the printed part is isotropic is problematic due to the manufacturing process of using fused deposition modeling. For future works this has to be addressed in a more accurate definition of the stiffness matrix.  Additionally, the assumption has been made, that the width of the platform is very small, which also doesn't accurately represent reality; the width is 40 mm and has therefore relatively same dimensions compared to the remaining dimensions of the RCC.

Additionally, the experiment itself can introduce error due to a rotation of the upper plane in between the gripper while applying a load in the $y$- direction. In this case the reference point would also be loaded with an additional moment, with which other elements of the stiffness matrix have to be considered in the analytical calculation.  The contact surface of the gripper should be increased in future experiments.

\section{Applications}
This section presents the application of the RCC devices for a variety of high-speed assembly tasks, videos of which are available here: \url{https://owncloud.fraunhofer.de/index.php/s/l7ZY2MpXQhRTUq2}

\subsection{Small RCC}
For the small RCC, two tasks have been defined, a peg in hole operation for gear assembly and switch rail assembly, as seen in Figure \ref{small_rcc}. The upper and bottom plane are connected using flexures with a 75° degree angle. The RCC serves to compensate positional and orientation errors in the assembly processes \cite{whitney2004}, and the larger range of motion ($\sim2$ cm) is used to allow a diagonal motion of the robot. The bottom plate is mounted to a linear positioning mechanism to allow a variation in the depth of the assembly using a worm gear. On the upper plane the connector can be mounted using the slotted hole.

\subsubsection{Gear assembly} 
For the connector a pin is used fitting the inner mounting diameter of a small gear as shown in Figure \ref{small_rcc_gear}. The RCC is important to prevent the gear jamming during the mount process due to tolerances in the rotation of the gear, which could lead to failure or unstable contact. The part is mounted diagonally from above, resulting in a final position as seen in  \ref{small_rcc_gear}, with RCC deformation of $2-3$ cm. Experiments have been performed with linear speeds up to 500 mm/sec. Even at full speed, $8$ mm tolerance in the $y$ direction and $4$ mm tolerance in the $z$ direction are achieved.

\subsubsection{DIN Rail Assembly}
\begin{figure}[h!]
	\centering
	\subfloat[Gear assembly with strong deformation
	\label{small_rcc_gear}]{\includegraphics[width=.55\columnwidth]{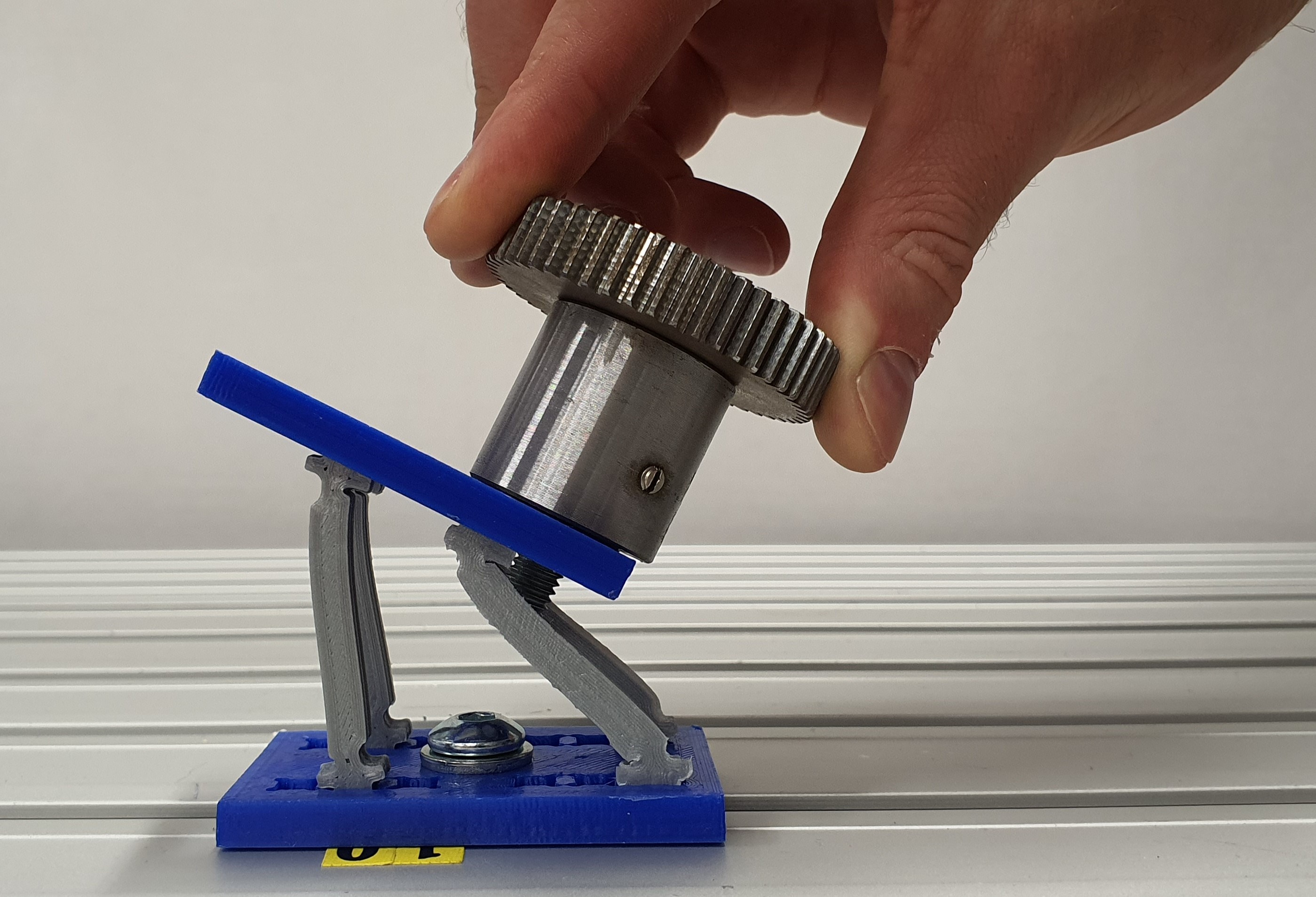}} 
	\hfill
	\subfloat[DIN rail switch assembly
	\label{assembled_rail}]{\includegraphics[width=.4\columnwidth]{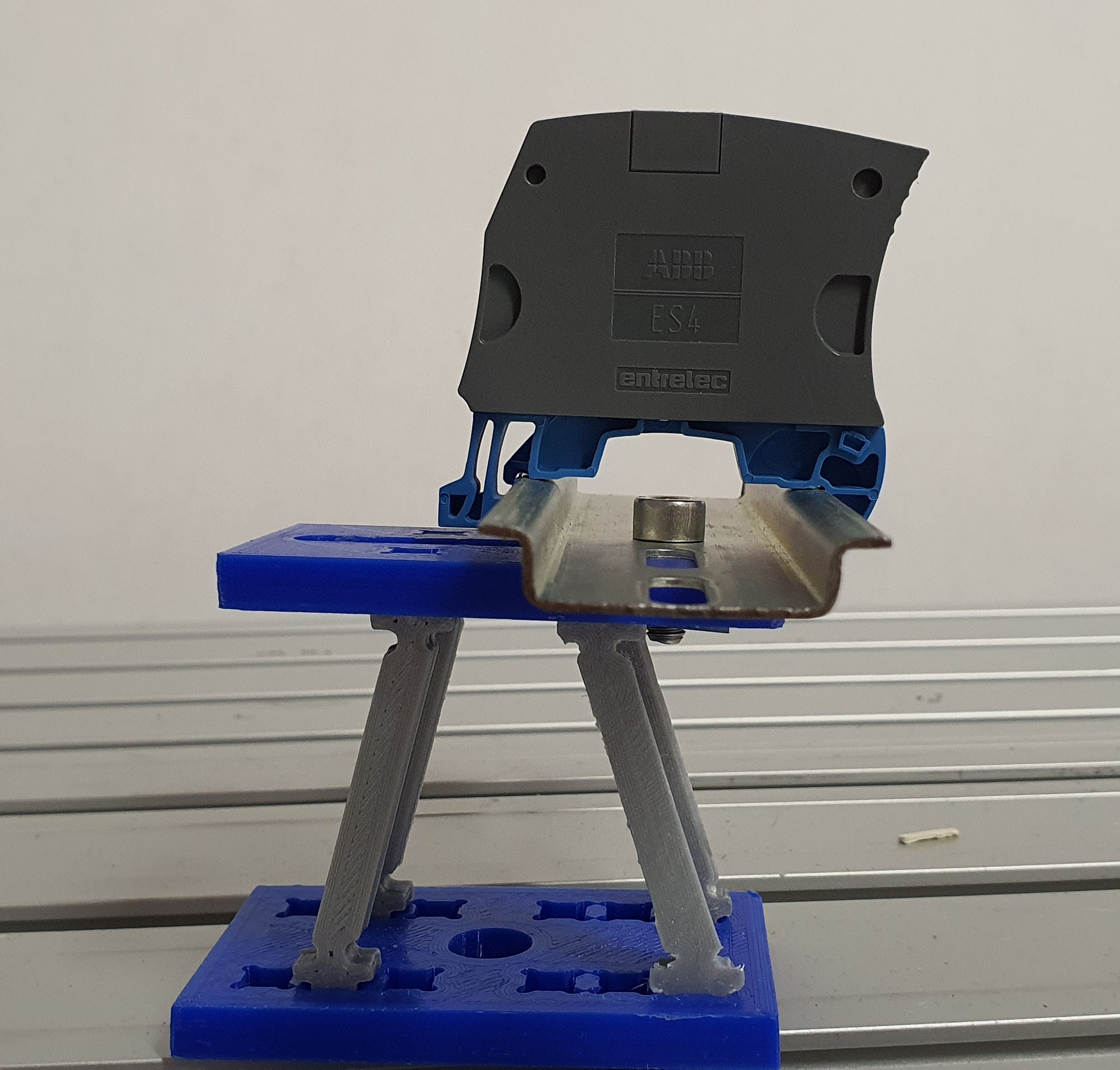}}
	\hfill
	\protect\caption{Two applications of the small RCC, where in (a) the approximate final pose of a gear assembly and (b) the assembled state of the rail assembly.\label{small_rcc}}
\end{figure}

\begin{figure}
\vspace{1mm}
    \centering
    \includegraphics[width=0.75\columnwidth]{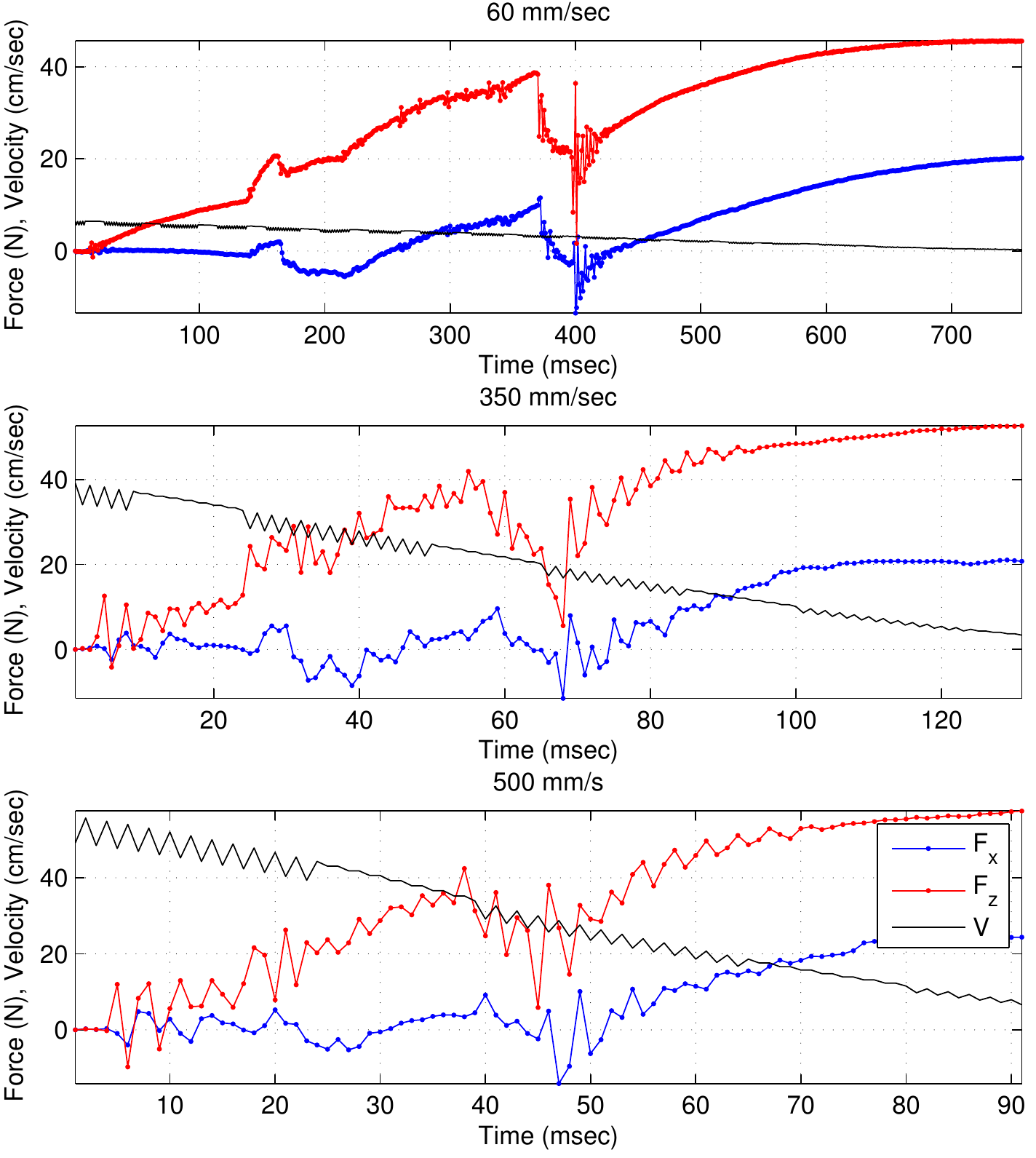}
    \caption{Assembly forces of the DIN Rail assembly at various speeds.  A low sprung inertia eliminates collision peaks, even at 500 mm/sec}
    \label{din_rail_time}
\end{figure}

A switch is mounted on a DIN rail. To ensure correct assembly first the lower right lip needs to clip into the rail, so a diagonal approach from the top right is again employed. In the next step the switch pushes the rail further down, causing a rotation due to the RCC, which allows the rail to clip into the upper lip, ensuring a correct assembly, as seen in \ref{assembled_rail}. This process ensures a robustness in 1.5 mm position variation at speeds up to $500$ mm/sec. The forces in assembly at various speeds can be seen in Figure \ref{din_rail_time}, which are time adjusted to align them by process stage.  The low sprung inertia keeps the peak collision force small, even at substantially higher speeds.

\subsection{Medium RCC / Large RCC}
\subsubsection{Peg-in-hole}
The PLA joints in Figure \ref{large_scale_versions} are used with standard aluminum profile to build a compliant table with rough top dimensions of $50\times60$ cm, as seen in \ref{large_scale}. Two pins on top are used for a dual peg-in-hole application, where a plate is guided into contact with the pins by manual guidance.   

\subsubsection{Plate Magazine}
\label{sec:storage_magazine}
\begin{figure}[h!]
	\centering
	{\includegraphics[width=.7\columnwidth]{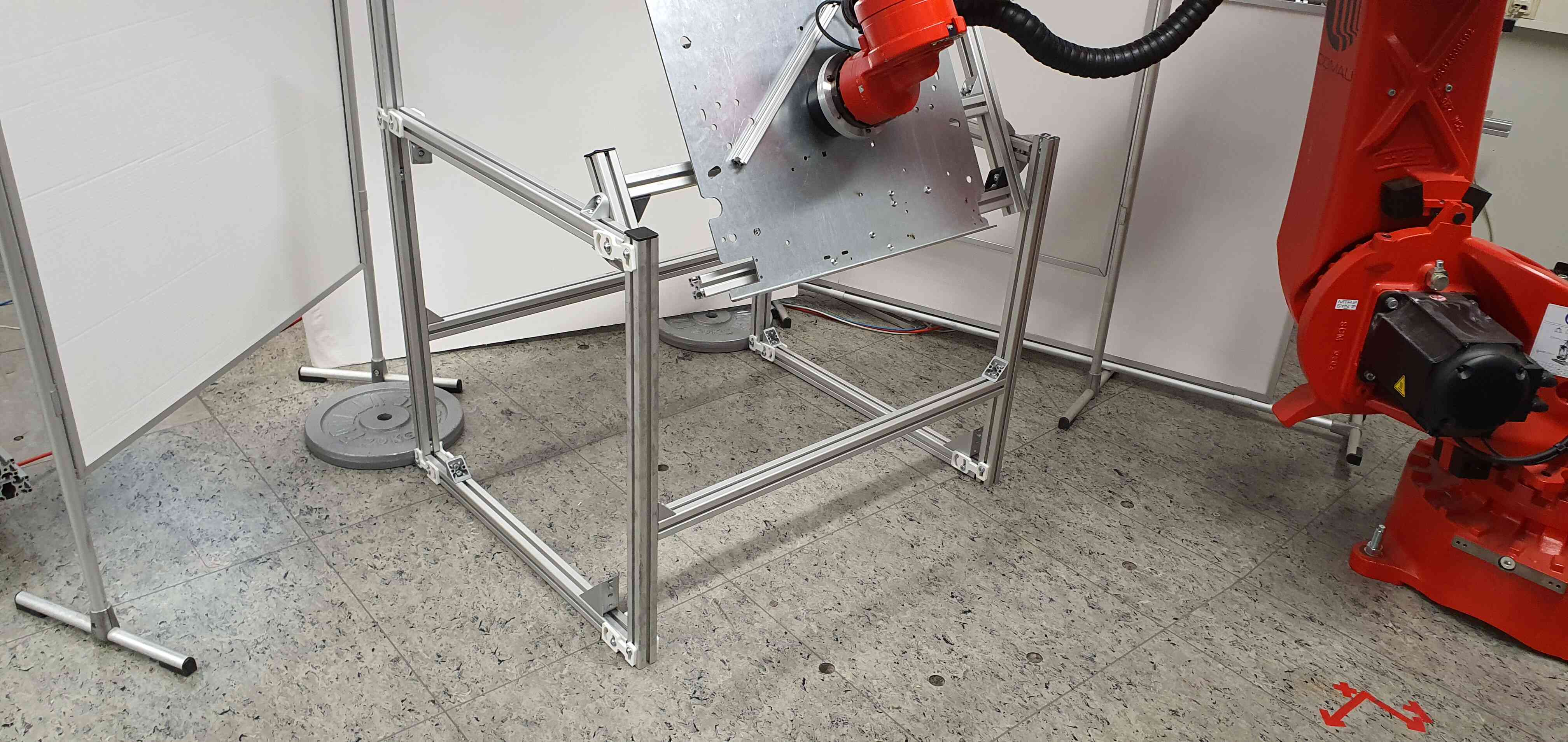}}
	\protect\caption{A large-scale compliant magazine for the mounting of plates.}
	\label{single_magazine}
\end{figure}

Thicker TPLA flexures ($17.5\times7$ mm for the narrowing) are used with aluminum to add compliance to a storage magazine for plates as shown in Figure \ref{single_magazine}. This magazine allows for higher-speed contact when the magazine is in an uncertain position.  When position uncertainty is larger, contact detection or force control can also be used.

\subsection{Vertical compliance}
Vertical compliance in only the z direction is realized with compliant materials and supporting rigid materials, as shown in \ref{z_direction} and \ref{z_direction_assembly}. To adjust the compliance, viscoelastic material can be exchanged or layered. Mounting screws can lock the vertical compliance to allow a fully stiff system. Four vertical compliant systems are integrated into the table seen in \ref{large_scale}. 

A major disadvantage of the current design is possible jamming of the aluminum profiles due to low tolerances between profile and slot in the linear guidance. The low tolerances are currently needed to ensure the only DOF being the movement in z direction. For future designs the linear guidance has to be revised to ensure the same DOF without causing the profiles to jam into the walls of the guidance. Nevertheless, the current design is applicable for a feasibility study of the work principle.

\section*{Conclusion}
Our contribution is the development of a low cost, easy to manufacture, highly adaptive and therefore easy to integrate solution for an RCC in the working space without further limiting the robot's payload and accuracy compared to the presented state-of-the-art solutions, allowing the robot design to remain unchanged whereas the compliant environment can easily be adapted to specific mounting tasks of different scale.  Further, this approach is shown to be feasible across a range of dimension and weight, allowing safe contact rich tasks, reducing contact forces, improving force control contact stability, and in some cases improving task robustness. However, making the environment compliant comes with trade offs, where additional design and integration effort is required, which can be streamlined due to our design's nature of being easily adjustable, iterative and easily integrated into industrial environments. The 6-DOF stiffness matrix of the small RCC has also been analytically determined and validated, developing tools for design of similar flexures. 

\bibliographystyle{IEEEtran}
\bibliography{bibliography/lib} 

\clearpage

\end{document}